\documentclass[letterpaper,10pt,conference]{ieeeconf}
\IEEEoverridecommandlockouts
\usepackage[letterpaper,top=56bp,bottom=76bp,left=48bp,right=48bp]{geometry}
\usepackage[T1]{fontenc}
\usepackage{times,amsmath,amssymb,graphicx,booktabs,array,url}
\usepackage[table]{xcolor}
\usepackage{stfloats}
\usepackage{float}
\usepackage[normalem]{ulem}
\definecolor{RevisionDeleted}{HTML}{C00000}
\definecolor{RevisionAdded}{HTML}{0057B8}

\usepackage[hidelinks]{hyperref}
\hypersetup{pdftitle={ExtremeDrive: A Real-World Dataset and Benchmark for Critical Driving},
pdfauthor={Xinhan Li; Yuhong Jiang; Shiyue Zhao; Junzhi Zhang; Chengkun He; Ruihai Ma; Shijie Zhao; Wenxuan Cheng; Jinbo Lv; Tao Liu}}
\title{ExtremeDrive: A Real-World Dataset and Benchmark for Critical Driving}
\author{\authorblockN{Xinhan Li$^{a,\dagger}$, Yuhong Jiang$^{a,\dagger}$, Shiyue Zhao$^{a}$, Junzhi Zhang$^{a,b,*}$, Chengkun He$^{a,b}$,\\
Ruihai Ma$^{a,b}$, Shijie Zhao$^{a,b}$, Wenxuan Cheng$^{a}$, Jinbo Lv$^{a}$, Tao Liu$^{c}$}
\authorblockA{$^{a}$School of Vehicle and Mobility, Tsinghua University, Beijing, China\\
$^{b}$State Key Laboratory of Intelligent Green Vehicle and Mobility, Tsinghua University, Beijing, China\\
$^{c}$Xibu Zhiche (Chongqing) Technology Co., Ltd., Chongqing, China\\
$^{\dagger}$These authors contributed equally.\quad $^{*}$Corresponding author: \href{mailto:jzhzhang@mail.tsinghua.edu.cn}{jzhzhang@mail.tsinghua.edu.cn}}}
\makeatletter
\IEEEaftertitletext{%
\begin{minipage}{\textwidth}
\centering
\includegraphics[width=0.94\textwidth,trim=0 28bp 0 40bp,clip]{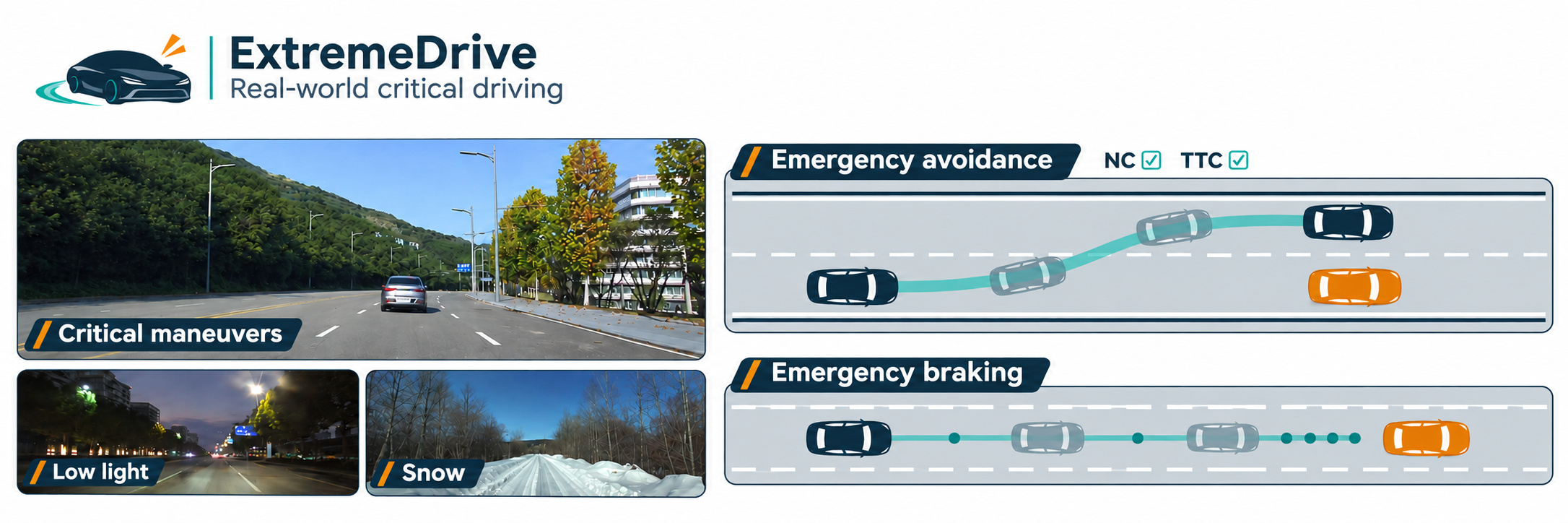}
\def\@captype{figure}
\caption{Overview of the ExtremeDrive dataset.}
\label{fig:overview}
\end{minipage}\vspace{0.2\baselineskip}
}
\makeatother
\begin{document}
\makeatletter\immediate\write\@auxout{\string\citation{BSTcontrol}}\makeatother
\maketitle
\thispagestyle{empty}\pagestyle{empty}

\begin{abstract}
Existing autonomous driving datasets and benchmarks provide extensive resources for routine driving, yet real-world data covering safety-critical scenarios and extreme environmental conditions remain limited.
To address this gap, we present ExtremeDrive\footnote{Toolkit and evaluation assets: \url{https://github.com/XinhanLi21/ExtremeDrive}}, a real-world dataset and benchmark covering emergency maneuvers, complex traffic interactions, and adverse conditions such as slippery or snow-covered roads, alongside routine driving.
The dataset comprises 589 driving scenes curated from a larger pool of real-world driving recordings, with calibrated and aligned multisensor data and reviewed annotations of road structure and traffic participants.
We further establish a benchmark to evaluate planned trajectories through simulated vehicle execution, measuring safety-related metrics and driving performance.
Experiments at equal training set sizes show that incorporating critical driving data reduces trajectory prediction errors and improves mean critical-success rates in critical scenarios.
ExtremeDrive provides a unified resource for learning and evaluating autonomous driving in safety-critical scenarios and extreme environmental conditions.
\end{abstract}

\begin{figure*}[!t]\centering
\includegraphics[width=\textwidth]{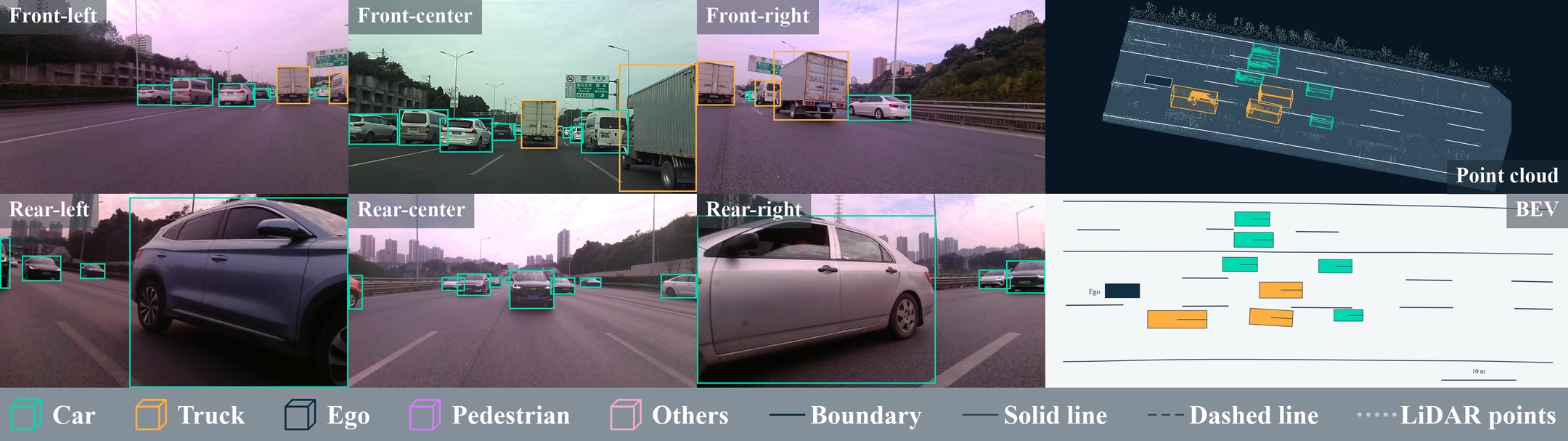}
\caption{Multisensor views of a complex traffic scene in ExtremeDrive. Camera detections accompany a LiDAR reconstruction and a BEV view of road and traffic annotations.}
\label{fig:multisensor}\end{figure*}

\section{Introduction}
Autonomous driving datasets and benchmarks provide observations and evaluation tools for developing and assessing driving systems~\cite{nuscenes,nuplan}.
However, most datasets commonly used for autonomous driving focus on routine driving scenarios and basic driving maneuvers, with particularly limited coverage of emergency maneuvers and extreme environmental conditions~\cite{wode2e,acdc}.
This imbalance leaves fewer opportunities to learn from and evaluate driving behavior under critical conditions.

Rare but consequential driving events are central to the long-tail challenge of autonomous driving: their low frequency makes them difficult to cover, while failures can have serious safety consequences~\cite{safebench}.
Complex traffic interactions, low light, wet or slippery roads, and snow-covered roads introduce additional challenges in perception, decision making, and vehicle response.
Strong performance under routine driving conditions does not establish planning capability in critical scenarios.
Real-world critical driving data are therefore needed to study these challenges and support the development of safer driving systems.
To support learning and evaluation, these data must also be consistently processed and supplemented with reliable annotations of road structure and traffic participants~\cite{interaction}.

Building on these data and annotations, assessing the ability to respond safely in critical scenarios requires suitable evaluation criteria.
Trajectory accuracy measures agreement with the recorded trajectory, but does not by itself establish whether a planned response can be executed safely~\cite{navsim,pseudosim}.
Different responses may also be safe, so reproducing a single recorded trajectory should not be the sole criterion for success.
Vehicle response and surrounding road and traffic constraints must be considered when assessing executed outcomes~\cite{paden}.
These needs motivate combining real-world critical driving data with evaluation that accounts for both trajectory accuracy and execution-based safety metrics.

To address these challenges, we introduce ExtremeDrive, a real-world dataset and benchmark for critical driving (Fig.~\ref{fig:overview}).
It covers critical scenarios involving complex traffic interactions and adverse conditions, alongside routine driving conditions.
The dataset provides calibrated and aligned multisensor data, vehicle states, and reviewed scene annotations from recordings collected on real roads.
The accompanying benchmark evaluates planned trajectories through simulated vehicle execution, measuring driving performance and safety-related metrics in critical scenarios.

Our contributions are threefold:

\smallskip
\noindent(1) \textbf{We construct ExtremeDrive, a real-world dataset for critical driving,} from curated recordings collected with an instrumented vehicle on real roads.
It covers emergency maneuvers during hazardous traffic interactions, including evasive steering and emergency braking, as well as extreme driving environments such as slippery roads and roads covered by snow or ice.
The dataset provides calibrated and aligned multisensor observations with annotations of the traffic environment across complete scenes.

\smallskip
\noindent(2) \textbf{We establish a unified benchmark for comparing planning performance under shared vehicle execution settings}.
It assesses planned trajectories through a shared tracking and vehicle execution interface, measuring collision avoidance, road compliance, progress, and comfort, with a dedicated critical-success criterion.

\smallskip
\noindent(3) \textbf{We validate the value of critical driving data and benchmark representative planning methods.}
Controlled experiments quantify the prediction benefit of critical driving data at equal training set sizes, while comparisons of representative planning methods characterize trajectory accuracy and execution performance.

\section{Related Work}

\subsection{Datasets for Challenging Driving}
Datasets for adverse conditions examine how weather, visibility, and seasonal changes affect perception and localization.
Seeing Through Fog studies weather-dependent sensor degradation~\cite{seeingfog}; ACDC pairs adverse and normal views with uncertainty annotations~\cite{acdc}.
Boreas captures seasonal variation~\cite{boreas}, while Boreas-RT extends coverage to challenging road environments for odometry and localization~\cite{boreasrt}.

Another group of datasets addresses hazardous events, complex interactions, and long-tail scenarios, with annotations tailored to different tasks.
DoTA labels when, where, and what traffic anomalies occur~\cite{dota}, while CODA benchmarks corner case detection~\cite{coda}.
INTERACTION captures negotiations, merging, and near-collision behavior through trajectories and semantic maps~\cite{interaction}.
WOD-E2E combines curated long-tail driving scenarios with trajectory evaluation based on human preference labels~\cite{wode2e}.
ExtremeDrive complements these resources by connecting real recordings of critical driving scenarios with traffic environment annotations and aligned vehicle states, supporting the evaluation of vehicle responses to critical situations.

\subsection{Planning Benchmarks}
Planning benchmarks distinguish trajectory imitation from the behavior produced by a plan~\cite{parting}.
nuPlan combines real recordings with closed-loop simulation~\cite{nuplan}, while Waymax uses real driving data to initialize simulations with models of the behavior of surrounding traffic participants~\cite{waymax}.
In contrast, NAVSIM uses short-horizon nonreactive simulation without sequential sensor generation~\cite{navsim}.
Pseudo-simulation evaluates recovery from deviations through synthetic observations around possible future states~\cite{pseudosim}.
These approaches differ in how they update observations and surrounding traffic as the evaluated vehicle departs from its recorded trajectory.

Other benchmarks organize interactive evaluation around safety and specific driving abilities.
SafeBench organizes safety-critical scenario generation and evaluation~\cite{safebench}, while Bench2Drive assesses driving abilities through short interactive routes~\cite{bench2drive}.
Scenario selection, interaction design, and success criteria determine which aspects of driving capability these benchmarks assess.
Building on existing planning evaluation protocols, ExtremeDrive evaluates planned responses to recorded real-world critical events through vehicle execution under nonreactive traffic assumptions.

\section{ExtremeDrive Dataset}

\begin{table*}[t]
\centering
\caption{Comparison of selected driving datasets.}
\label{tab:datasets}
\footnotesize
\setlength{\tabcolsep}{3pt}
\renewcommand{\arraystretch}{1.06}
\newcommand{\dtcell}[1]{\begin{tabular}[c]{@{}c@{}}#1\end{tabular}}
\newcommand{\dthead}[1]{\makebox[0pt][c]{\rotatebox{58}{#1}}}
\begin{tabular}{>{\centering\arraybackslash}m{106pt}>{\centering\arraybackslash}m{101pt}*{8}{>{\centering\arraybackslash}m{29.475pt}}}
\toprule
Dataset & Scale & \dthead{Camera} & \dthead{LiDAR} & \dthead{Vehicle signals} & \dthead{Emergency} & \dthead{Urban} & \dthead{Suburban} & \dthead{Rural} & \dthead{Highway}\\
\midrule
\dtcell{nuPlan~\cite{nuplan}} & \dtcell{1,282 h / 4 cities} & $\checkmark$ & $\checkmark$ & --- & --- & $\checkmark$ & --- & --- & --- \\
\rowcolor{black!5}
\dtcell{INTERACTION~\cite{interaction}} & \dtcell{11 locations} & $\times$ & $\times$ & --- & $\checkmark$ & $\checkmark$ & --- & --- & $\checkmark$ \\
\dtcell{DoTA~\cite{dota}} & \dtcell{4,677 videos} & $\checkmark$ & $\times$ & --- & $\triangle$ & --- & --- & --- & --- \\
\rowcolor{black!5}
\dtcell{KITTI (Odometry)~\cite{kitti}} & \dtcell{39 km / 22 seqs} & $\checkmark$ & $\checkmark$ & --- & --- & $\checkmark$ & $\checkmark$ & $\times$ & $\times$ \\
\dtcell{UrbanLoco~\cite{urbanloco}} & \dtcell{40 km / 13 seqs} & $\checkmark$ & $\checkmark$ & --- & --- & $\checkmark$ & $\times$ & $\times$ & $\times$ \\
\rowcolor{black!5}
\dtcell{MulRan~\cite{mulran}} & \dtcell{124 km / 12 seqs} & $\times$ & $\checkmark$ & --- & --- & $\checkmark$ & --- & --- & --- \\
\dtcell{nuScenes~\cite{nuscenes}} & \dtcell{1,000 scenes / 20 s each} & $\checkmark$ & $\checkmark$ & $\checkmark$ & --- & $\checkmark$ & --- & --- & --- \\
\rowcolor{black!5}
\dtcell{Seeing Through Fog~\cite{seeingfog}} & \dtcell{12,000 road samples} & $\checkmark$ & $\checkmark$ & --- & --- & $\checkmark$ & $\checkmark$ & --- & $\checkmark$ \\
\dtcell{Boreas~\cite{boreas}} & \dtcell{350 km / 44 seqs} & $\checkmark$ & $\checkmark$ & $\checkmark$ & --- & $\checkmark$ & $\checkmark$ & $\times$ & $\times$ \\
\midrule
\rowcolor{black!5}
\textbf{ExtremeDrive (ours)} & \dtcell{\textbf{420 km / 589 scenes}} & $\checkmark$ & $\checkmark$ & $\checkmark$ & $\checkmark$ & $\checkmark$ & $\checkmark$ & $\checkmark$ & $\checkmark$ \\
\bottomrule
\end{tabular}
\par\smallskip
\begin{minipage}{\textwidth}\footnotesize
$\checkmark$: included; $\times$: absent; ---: not explicitly reported.
Vehicle signals: steering, torque, or wheel/encoder measurements.
Emergency denotes emergency maneuvers: recorded emergency braking or evasive maneuvers, irrespective of dedicated action labels.
In this column, $\triangle$ denotes reported braking-to-avoid examples without an explicit emergency-severity designation; --- denotes unconfirmed coverage, not confirmed absence.
\end{minipage}
\end{table*}

\subsection{Data Acquisition and Sensor Alignment}
\begin{figure}[!htb]
\centering
\includegraphics[width=\columnwidth,trim=0 0 0 230bp,clip]{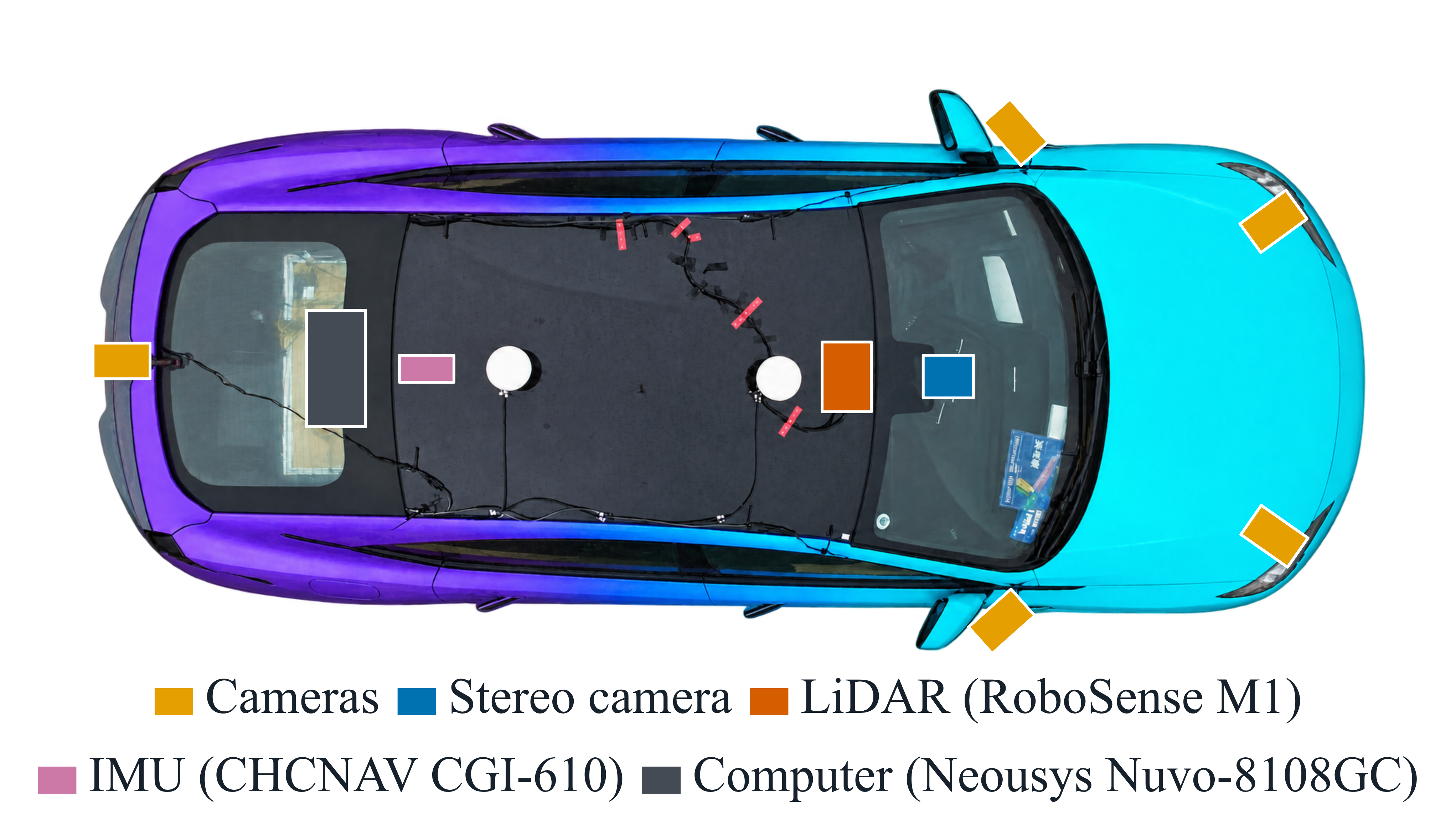}
\caption{ExtremeDrive data collection platform and sensor layout.}
\label{fig:vehicle}
\end{figure}

ExtremeDrive targets two complementary aspects of critical driving: emergency maneuvers in hazardous traffic interactions and extreme driving environments such as slippery roads and roads covered by snow or ice.
Data were collected in Chongqing and Mohe, China, using an instrumented Changan Deepal SL03 (Fig.~\ref{fig:vehicle}; vehicle parameters in Table~\ref{tab:vehicle_parameters}).

The acquisition platform combines a stereo camera at $1280\times720$ pixels, five cameras at $640\times480$ pixels, and a LiDAR with a $120^{\circ}$ horizontal field of view.
An integrated GNSS/IMU unit with RTK capability records vehicle position and inertial motion.
The vehicle CAN FD interface provides information on vehicle dynamics, driver inputs, and actuator responses.

\begin{table}[!htb]
\centering
\caption{Key vehicle parameters of the Changan Deepal SL03 platform.}
\label{tab:vehicle_parameters}
\footnotesize
\setlength{\tabcolsep}{4pt}
\begin{tabular}{lcl}
\toprule
\textbf{Parameter} & \textbf{Symbol} & \textbf{Value} \\
\midrule
Total mass & $m$ & $2178\,\mathrm{kg}$ \\
Yaw moment of inertia & $I_z$ & $3216\,\mathrm{kg}\cdot\mathrm{m}^2$ \\
CG-to-front-axle distance & $l_f$ & $1.526\,\mathrm{m}$ \\
CG-to-rear-axle distance & $l_r$ & $1.374\,\mathrm{m}$ \\
Wheel rotational inertia & $I_w$ & $2.5\,\mathrm{kg}\cdot\mathrm{m}^2$ \\
\bottomrule
\end{tabular}
\end{table}

Multisensor calibration and temporal alignment associate observations of the surrounding scene with the recorded vehicle state.
An onboard computer runs ROS acquisition nodes and stores timestamped vehicle signals linked to camera images and LiDAR arrays, supporting temporal alignment during processing.
The corresponding vehicle poses link observations in the ego frame to annotations in the scene frame.
The selected records are provided at a nominal rate of 4\,Hz, with camera, LiDAR, IMU/GNSS, and vehicle state information.
Table~\ref{tab:datasets} compares ExtremeDrive with existing driving datasets and benchmarks in terms of sensor modalities, vehicle signals, emergency maneuver coverage, and road coverage.

\begin{figure*}[t]\centering
\includegraphics[width=\textwidth,trim=0 10bp 0 4bp,clip]{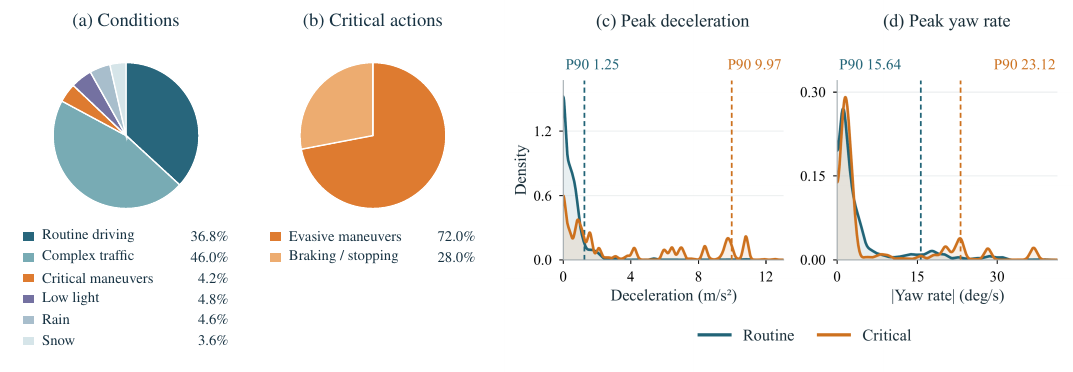}
\caption{Dataset composition and maneuver intensity. (a) Condition shares among 589 driving scenes. (b) Action shares among critical scenes. Percentages are rounded. (c)--(d) Normalized peak-deceleration and absolute yaw rate densities for routine and critical clips over 4\,s horizons; dashed lines mark empirical P90 values. Each density panel uses a shared bandwidth and reflection at zero.}
\label{fig:difficulty}\end{figure*}

\subsection{Dataset Composition and Scenario Coverage}
ExtremeDrive comprises 589 driving scenes curated from approximately 20\,h of raw recordings, with 420\,km of acquisition mileage (rounded).
The annotation layer provides complete scene maps.
Each driving scene contains a continuous sequence of recorded observations and vehicle states, from which planning clips are extracted.

\textbf{Scenario coverage.}
The recordings cover emergency braking and evasive maneuvers in hazardous interactions, as well as driving on slippery roads and roads covered by snow or ice.
Emergency braking and evasive maneuvers were deliberately organized as real-vehicle tests under controlled safety conditions; their labels reflect the executed driving actions and maneuver intensity.
Additional recordings cover routine driving conditions, complex traffic interactions, low light, and rain.
Fig.~\ref{fig:difficulty}(a)--(b) summarizes condition shares among the 589 driving scenes and action shares among its critical scenes.
Evasive maneuvers account for 72.0\% of these critical scenes, while braking or stopping accounts for 28.0\%.

\textbf{Maneuver intensity.}
Criticality is reflected in stronger braking and steering demands, rather than a higher median initial speed.
Median anchor speeds are 10.55 and 10.56\,m/s for routine and critical clips, respectively.
The motion analysis uses routine and critical clips, with peak deceleration and peak absolute yaw rate evaluated over 4\,s horizons.
Fig.~\ref{fig:difficulty}(c)--(d) shows that critical P90 peak deceleration is nearly eight times the routine value (9.97 versus 1.25\,m/s$^2$).
P90 peak absolute yaw rate is also higher (23.12 versus 15.64\,deg/s).

\textbf{Planning samples and splits.}
Each planning clip starts at a recorded timestamp and includes the 4\,s future ego trajectory; neighboring clip start times are spaced at 0.25\,s where data support permits.
The planning clips are divided into training, validation, and test sets.
All clips from the same source scene belong to a single partition, and annotation map identifiers are not shared across partitions.

The critical subset is defined by emergency maneuver labels.
The planning split includes critical source scenes in the training and test sets.

\subsection{Scene Annotations and Causal Descriptions}
The annotation layers connect traffic environment geometry, scene semantics, and descriptions of driving decisions.

\textbf{Complete scene geometry.}
Image evidence initializes the appearance and topology of road markings, while LiDAR geometry and reflectance constrain their position and physical boundaries.
Editable curves and polygons represent road markings, drivable surfaces, and physical exclusions.
Linked camera and BEV views support human correction of continuity, curvature, dashed gaps, and drivable surfaces across a complete scene.
Camera--LiDAR association and temporal tracking provide three-dimensional reference annotations for traffic participants and tracks with consistent identities.

The drivable surface follows reviewed outer markings and the driven road corridor, excluding annotated physical obstacles.
The resulting scene representations combine continuous road geometry, drivable areas, and consistent traffic participant trajectories.
Together with vehicle poses, these annotations provide scene constraints for planning and geometric references for evaluating alternative responses.

\textbf{Semantic and motion annotations.}
Scene-level semantic annotations describe weather, traffic, road surface, driving intent, hazards, and criticality, together with scene summaries and confidence fields.
A vision-language model (VLM) assists the labeling of these attributes.
Per-frame metadata pairs vehicle states with relative position and heading increments, providing motion supervision alongside scene descriptions.

\begin{table*}[!t]\centering\small
\renewcommand{\arraystretch}{1.22}
\caption{Planning and execution results.}
\label{tab:models}
\setlength{\tabcolsep}{3.4pt}
\begin{tabular}{llrrrrrrrr}\toprule
Method & Input & ADE/FDE (m)$\downarrow$ & $S^+\uparrow$ & Collision$\downarrow$ & Road$\uparrow$ & TTC$\uparrow$ & Progress$\uparrow$ & $C\uparrow$ & ES$\uparrow$\\\midrule
HyperDiffusion~\cite{hyperdiffusion} & V & 1.17/2.79 & 80.57 & 2.53 & 85.85 & 88.88 & 92.82 & 99.82 & 61.70\\
SimLingo~\cite{simlingo} & V & 1.47/3.44 & 77.58 & 3.97 & 82.19 & 87.95 & 90.20 & 99.78 & 51.77\\
Latent TransFuser~\cite{transfuser} & V & 1.18/2.91 & 75.46 & 3.38 & 80.33 & 88.45 & 92.05 & 99.90 & 58.16\\
DiffusionDrive~\cite{diffusiondrive} & V+L & 1.28/3.33 & 74.85 & 3.30 & 79.86 & 88.65 & 92.51 & 96.88 & 57.45\\
TransFuser~\cite{transfuser} & V+L & 1.09/2.71 & 74.26 & 3.46 & 77.95 & 88.57 & 92.60 & 99.93 & 56.03\\
\bottomrule\end{tabular}
\par\smallskip
\begin{minipage}{\textwidth}\footnotesize
V: camera; L: LiDAR; ES: critical success rate.
$S^+$: composite upper bound (0--100); rates, progress, and noncritical comfort $C$ are percentages over evaluable clips.
\end{minipage}
\end{table*}

\begin{figure}[H]\centering
\includegraphics[width=\columnwidth,trim=0 12bp 0 8bp,clip]{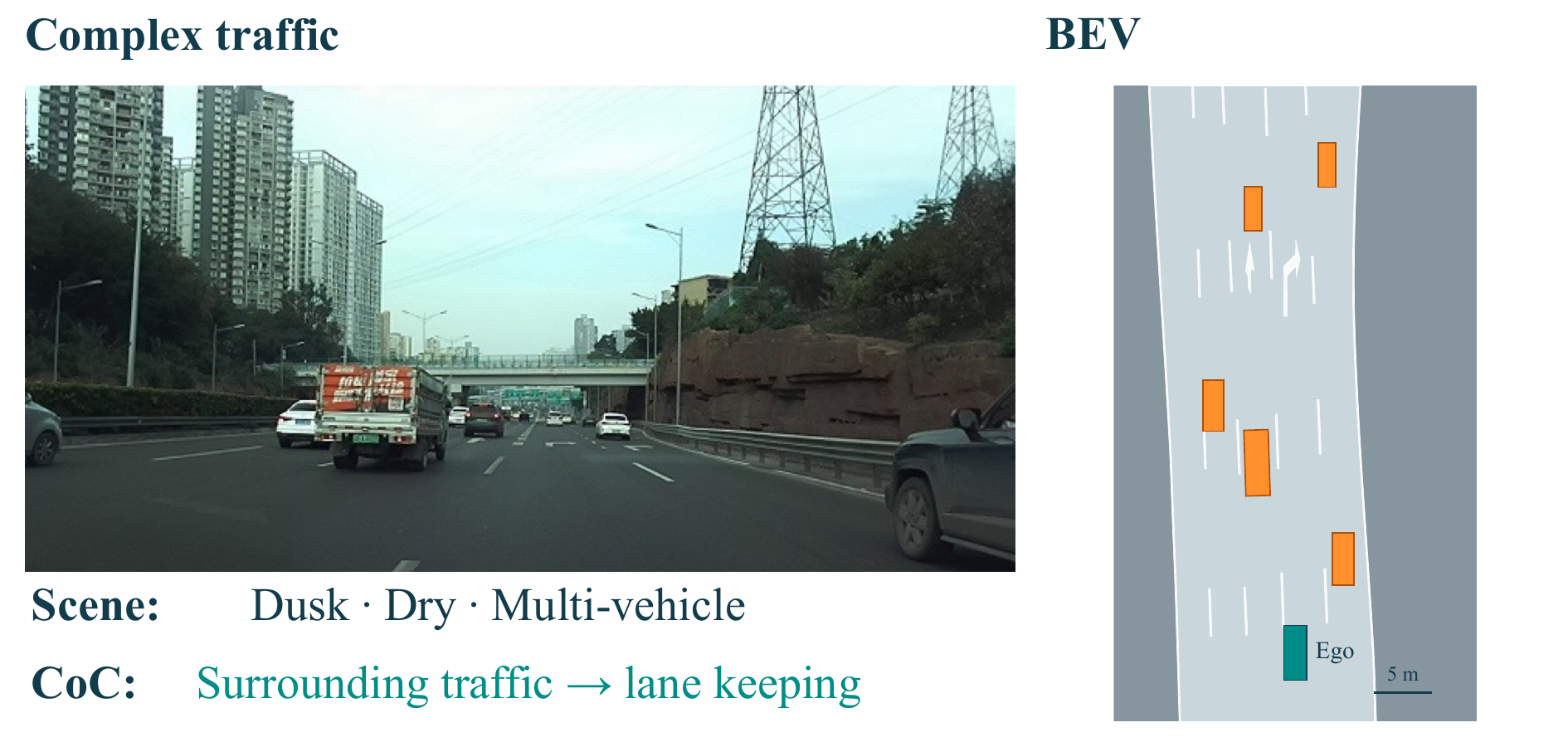}
\caption{Annotation example from a multi-vehicle scene. Scene labels and CoC summarize the original annotations; the BEV shows accepted road annotations and in-road vehicle footprints aligned with the camera frame. Shading distinguishes the exterior of the solid boundaries.}
\label{fig:annotations}\end{figure}

\textbf{Causal descriptions.}
The dataset also includes Chain of Causation (CoC) records.
They connect scene evidence, critical factors, driving decisions, and longitudinal/lateral meta-actions.
These causal annotations are generated with a VLM to provide auxiliary causal supervision.
The VLM-generated annotations were manually verified.
Fig.~\ref{fig:annotations} illustrates these annotation layers.

\textbf{Availability and privacy.} The dataset is hosted on \href{https://huggingface.co/datasets/Stary108/Extreme_Driving_Conditions_Dataset}{Hugging Face} under the CC BY-NC 4.0 license. Faces and vehicle license plates in the released RGB images are blurred to protect privacy.

\section{ExtremeDrive Benchmark}
\subsection{Evaluation Protocol}
ExtremeDrive evaluates planned responses through simulated trajectory execution within recorded road and traffic contexts.
At each recorded anchor, a planner uses its designated sensor inputs to predict eight future ego poses at 0.5\,s intervals.
A shared tracker and vehicle model execute this trajectory from the recorded vehicle state.
Surrounding traffic participants follow their recorded trajectories; the planner supplies one reference over the 4\,s horizon~\cite{navsim}.
Driving and safety metrics are then computed from the executed motion and the annotated traffic environment.

Feedback is confined to ego trajectory tracking, with a fixed planned trajectory and recorded surrounding traffic.

Evaluation uses reviewed maps and temporally aligned traffic participant tracks.
A shared set of traffic participants is used to assess risk for every planner.

\subsection{Trajectory Execution}
During emergency maneuvers, vehicle response affects the executed path and its collision and road compliance outcomes.
We execute planned trajectories through a shared tracker and a seven-degree-of-freedom (7DOF) vehicle dynamics model~\cite{hou2022crash}, with parameters shown in Table~\ref{tab:vehicle_parameters}.
The predicted poses are transformed from the anchor ego frame into the scene frame and interpolated onto a 10\,Hz reference.
A shared linear quadratic regulator (LQR) tracker~\cite{navsimcode} converts trajectory tracking errors into acceleration and steering rate requests.
Vehicle states are expressed using the rear-axle convention required by the tracker, and simulated motion and steering provide feedback during trajectory tracking.

The model couples planar vehicle motion with four wheel rotations.
Wheel forces use a slip-based approximation with a smooth combined-force limit~\cite{rajamani}.
Road friction is assigned empirically from scene-level surface conditions ($\mu=0.30$, $0.60$, and $0.85$ for snow, wet, and dry roads, respectively) and held fixed across planners evaluated in the same scene.
Control requests are subject to steering and torque limits, actuator delay, and first-order response before numerical integration at 1000\,Hz.
Vehicle parameters are fixed across planners, and metrics are computed from the executed vehicle footprint.
The executed body polygon is checked against road geometry and traffic participant footprints, allowing tracking deviations and vehicle response to affect collision and road compliance outcomes for a fixed planned trajectory.

We compare the calibrated 7DOF model with a kinematic bicycle model through measured-input replay on six driving logs excluded from calibration.
Both models receive identical measured control inputs and initial states.
As shown in Table~\ref{tab:vehicle_replay}, the calibrated 7DOF model achieves lower RMSE in position, longitudinal velocity, and lateral velocity under both conditions.
The largest relative improvements occur in lateral velocity, with RMSE reductions of 32.2\% on dry roads and 22.7\% on wet roads.
These results support the use of the calibrated 7DOF model to represent vehicle motion in the benchmark.

\begin{table}[H]\centering\footnotesize
\caption{Vehicle-model replay errors on held-out driving logs. All metrics are RMSE ($\downarrow$).}
\label{tab:vehicle_replay}
\setlength{\tabcolsep}{2.5pt}
\renewcommand{\arraystretch}{1.10}
\begin{tabular}{llrrr}\toprule
Condition & Model & Position & Long. vel. & Lat. vel.\\
 & & (m) & (m/s) & (m/s)\\\midrule
Dry & Kinematic bicycle & 0.697 & 0.378 & 0.081\\
Dry & Calibrated 7DOF & \textbf{0.688} & \textbf{0.373} & \textbf{0.055}\\
\midrule
Wet & Kinematic bicycle & 0.717 & 0.428 & 0.058\\
Wet & Calibrated 7DOF & \textbf{0.698} & \textbf{0.415} & \textbf{0.045}\\
\bottomrule\end{tabular}
\end{table}

\subsection{Metrics and Aggregation}
Evaluation combines collision avoidance and road compliance with progress, comfort, and a dedicated success criterion for critical responses.

\textbf{Collision and road compliance.}
Let $B(t)$ denote the executed ego footprint, $O_j(t)$ a traffic participant footprint, $D$ the reviewed drivable surface, $X$ the physical exclusion layer, and $L$ the prohibited marking geometry.
The no-collision gate $N$ checks intersections between $B$ and the recorded footprints of traffic participants.
Road compliance $R$ is the conjunction of marking compliance, surface containment, and exclusion avoidance when that layer is present.
Solid white and yellow markings define prohibited crossing geometry, while dashed separators are permitted.
Surface containment uses a 0.15\,m outward tolerance.
Geometric evaluation runs at 20\,Hz and includes additional checks for crossing road markings.

\textbf{Anticipatory safety, progress, and comfort.}
TTC compliance $T$ uses constant-velocity forecasting~\cite{navsimcode}, with a 1\,s horizon and 0.1\,s increments.
Observations of traffic participants at neighboring timestamps provide velocity estimates, while current overlap establishes zero TTC independently of velocity.
Progress $P\in[0,1]$ measures advancement along the recorded route relative to its horizon length.
Route lengths below 1\,m use a stationary-reference rule; leaving the route by more than 8\,m gives zero progress.
Comfort $C$ aggregates acceleration, jerk, yaw rate, and yaw acceleration exceedances.
If $\rho(t)$ is the largest normalized limit ratio, then $C=H^{-1}\int_0^H\max(1,\rho(t))^{-2}dt$.
Thresholds follow the reference implementation~\cite{navsimcode}; the continuous mapping retains exceedance magnitude and duration.

\textbf{Composite score and critical success.}
Following safety-gated weighted aggregation~\cite{navsim}, the composite score is
\begin{equation}
S=100\,NR\,\frac{5P+5T+2aC}{10+2a},\label{eq:score}
\end{equation}
where $a=0$ for preassigned critical maneuvers and $a=1$ otherwise.
Comfort remains part of the score for adverse environmental conditions without a critical maneuver label.
The binary collision and road gates prevent progress from compensating for a failed safety constraint.

Critical success is $E=N R T_{0.5}$ over a critical clip, where $R$ denotes road compliance and $T_{0.5}$ applies the same TTC test with a 0.5\,s forecast.
The composite score retains the 1\,s term $T=T_{1.0}$: these terms describe imminent and anticipatory collision risk, respectively.
This criterion requires collision avoidance, road compliance, and short-horizon TTC compliance during critical responses.
Comfort is excluded from the critical success criterion.
Critical success is evaluated over the 4\,s horizon.

Reported $S^+$ values are upper bounds for the composite score.
Other rates, progress, and noncritical comfort are aggregated over clips evaluable for the corresponding metric; critical success is reported on critical clips.

\section{Experiments}

\subsection{Experimental Setup}
Two experiments examine the training value of critical data and planning performance under the benchmark.
Trajectory accuracy is measured by average displacement error (ADE) and final displacement error (FDE) over the 4\,s horizon.
The first isolates the contribution of critical training data using TransFuser and DiffusionDrive, with a frozen-feature MLP as a lightweight baseline.
The MLP combines frozen image features with ego state variables to predict the future trajectory.
Perception encoders remain frozen, and all parameters in each planner's planning network are trained from the same pretrained initialization across the paired conditions.
Each arm contains 3,161 clips; the mixed arm replaces 177 routine clips with critical examples.
We use three paired random seeds and 100 training epochs, keeping optimization settings fixed within each planner and selecting checkpoints by ADE on the unchanged validation split.
Uncertainty is estimated by bootstrap resampling at the source scene level.

The second experiment evaluates adapted representative planning methods on the shared test split and execution protocol.
For sensor-based planners, visual encoders remain frozen while planning components are trained.
Inputs comprise cameras or cameras with LiDAR, as specified in Table~\ref{tab:models}.

\subsection{Critical Scenario Analysis}
\textbf{Critical examples improve prediction and mean critical success.}
At equal training sizes, adding critical data reduces critical ADE by 29.5\% for TransFuser and 33.0\% for DiffusionDrive; FDE decreases by 33.2\% and 36.0\%, respectively (Table~\ref{tab:benefit}).
Both planners improve critical ADE and critical success in all three paired runs.
Source-scene bootstrap 95\% intervals for the seed-averaged ADE change are $[-3.07,-0.24]$\,m and $[-3.67,-0.34]$\,m, respectively.

\begin{table}[t]\centering\small
\renewcommand{\arraystretch}{1.22}
\caption{Contribution of critical data at equal training size.}
\label{tab:benefit}
\setlength{\tabcolsep}{2pt}
\begin{tabular}{@{}lrrr@{}}\toprule
Model / training & \multicolumn{3}{c}{Critical}\\
 & ADE & FDE & ES (\%)\\\midrule
MLP-Routine only & 4.11$\pm$0.03 & 8.06$\pm$0.15 & 42.35$\pm$0.26\\
MLP-Mixed & 3.16$\pm$0.11 & 6.06$\pm$0.32 & 59.10$\pm$0.54\\
\midrule
TransFuser-Routine only & 3.28$\pm$0.09 & 8.28$\pm$0.20 & 47.43$\pm$0.20\\
TransFuser-Mixed & 2.31$\pm$0.16 & 5.53$\pm$0.51 & 57.45$\pm$2.33\\
\midrule
DiffusionDrive-Routine only & 3.53$\pm$0.11 & 8.72$\pm$0.22 & 46.35$\pm$1.55\\
DiffusionDrive-Mixed & 2.36$\pm$0.03 & 5.58$\pm$0.04 & 57.05$\pm$0.71\\
\bottomrule\end{tabular}
\par\smallskip
\begin{minipage}{\columnwidth}\footnotesize
ADE/FDE in meters; ES: critical success rate; $\pm$ denotes standard deviation.
\end{minipage}
\end{table}

Mean critical success rises from 47.43\% to 57.45\% for TransFuser and from 46.35\% to 57.05\% for DiffusionDrive.
These results show that incorporating critical driving data at a fixed training set size improves mean success rates in critical scenarios.

\subsection{Planner Comparison}
Table~\ref{tab:models} compares five adapted planners on the same test split using shared execution settings and scene-specific friction assignments.

Lower trajectory error does not necessarily yield higher execution-based safety scores.
TransFuser achieves the lowest ADE/FDE (1.09/2.71\,m), but HyperDiffusion achieves higher road compliance (85.85\% versus 77.95\%) and critical success (61.70\% versus 56.03\%), with ADE/FDE of 1.17/2.79\,m.
HyperDiffusion also attains the highest composite upper bound (80.57) and the lowest collision rate (2.53\%) among the evaluated methods.
Aggregate performance and critical success can also rank planners differently: SimLingo scores higher than Latent TransFuser overall (77.58 versus 75.46), while its critical success rate is lower (51.77\% versus 58.16\%).

\section{Discussion and Conclusion}
This work presents ExtremeDrive, a dataset and benchmark for critical driving.
Curated from vehicle recordings collected on real roads, the dataset covers emergency maneuvers and extreme driving environments.
Multisensor calibration, temporal alignment, and traffic environment annotations across complete scenes support learning and evaluation.
A unified benchmark assesses planned trajectories through a shared tracking and vehicle execution interface, combining driving performance metrics with a dedicated safety criterion for critical scenarios.
Equal-size experiments with TransFuser and DiffusionDrive show lower critical trajectory errors and higher mean critical success after incorporating critical data.
Comparisons of representative planners show different rankings across trajectory error, execution-based safety metrics, and aggregate scores.
ExtremeDrive provides data and evaluation tools for research on planning and control in critical driving scenarios.

Future work will prioritize increasing the dataset size by collecting more independent critical driving scenes across diverse road conditions and acquisition locations to support broader generalization studies.
Additional physical measurements can strengthen vehicle response validation during emergency maneuvers, particularly under low friction and high slip conditions.
Extending the benchmark to sequential replanning and reactive traffic participants will enable evaluation of sustained emergency responses and traffic interactions.
Further research will explore how annotations across complete scenes and causal descriptions can support learning-based planning in critical driving scenarios.

\bibliographystyle{IEEEtran}
\bibliography{references}
\end{document}